# Colorisation et texturation temps réel d'environnements urbains par système mobile avec scanner laser et caméra fish-eye

*Jean-Emmanuel Deschaud, Xavier Brun, François Goulette*

*Mines ParisTech, CAOR-Centre de Robotique, Mathématiques et Systèmes*
*60 Boulevard Saint-Michel 75272 Paris Cedex 06*
*E-mails : prenom.nom @mines-paristech.fr*

**Résumé**

Nous présentons ici un système temps réel de modélisation d'environnements urbains monté sur un véhicule. Le système d'acquisition terrestre est basé sur un système de géolocalisation et un ensemble de deux capteurs, un scanner laser et une caméra avec objectif fish-eye. Nous produisons des nuages de points 3D colorés et des modèles surfaciques triangulés et texturés de l'environnement traversé par le véhicule. Une fois que notre système a été calibré, l'acquisition et le traitement des données se fait "en vol". Cet article portera principalement sur nos différentes méthodes de colorisation des nuages de points, de triangulation et d'ajout de textures.

**Mots Clés :** système de cartographie mobile, nuage de points colorés, modèles texturés, scanner laser, fish-eye, caméra

*Abstract*

*We present here a real time mobile mapping system mounted on a vehicle. The terrestrial acquisition system is based on a geolocation system and two sensors, namely, a laser scanner and a camera with a fish-eye lens. We produce 3D colored points cloud and textured models of the environment. Once the system has been calibrated, the data acquisition and processing are done "on the way". This article mainly presents our methods of colorization of point cloud, triangulation and texture mapping.*

## Introduction

Les modèles d'environnements urbains sont utiles dans de nombreuses applications comme l'architecture, la cartographie 3D, la navigation terrestre, le tourisme virtuel, la réalité virtuelle, les jeux-vidéos, etc. De nombreuses équipes travaillent dans ce domaine de la construction automatique de bâtiments soit à partir de données images (stéréovision), soit à partir de données laser et image. Nous avons fait le choix de travailler à partir de données laser pour construire une géométrie plus précise. Nous recherchons une erreur moyenne de reconstruction inférieure à 5 cm, ce qu'il serait possible d'avoir à partir de données laser, alors que pour des modèles reconstruits à partir d'images l'erreur moyenne se situe dans l'état actuel de la recherche plutôt entre 5 et 40 cm [POL08].

A partir de notre système existant décrit dans [ANL05, GOU06, BRU07, BRU07b] qui permet la création de nuages de points 3D avec un scanner laser et un système de localisation, notre principale motivation a été d'ajouter de la couleur avec un nombre limité de caméras. Nous avons voulu nous placer dans un contexte temps réel pour la construction de modèles 3D : c'est-à-dire aucune perte de données lors du rejeu à 100% sur un PC de bureau. Ce choix du temps réel s'explique par la quantité de plus en plus importante de données brutes obtenues par tous les capteurs. Il nous semble possible de pouvoir créer une chaîne complète de traitement temps réel qui pourrait à terme être intégrée complètement sur la plate-forme mobile. Les modèles 3D pourraient alors être obtenus directement à la fin des acquisitions.

Cet article portera ainsi sur l'ajout de la couleur au nuage de points avec la création d'un nuage de points 3D coloré et de modèles à facettes texturés. Dans une première partie, nous donnerons un aperçu des différentes approches. Dans une seconde partie, nous exposerons rapidement le système existant pour expliquer ce qui a été fait pour l'améliorer avec l'ajout de la caméra fish-eye. Nous exposerons aussi dans cette même partie les processus mis en place pour faire l'étalonnage du scanner laser et de la caméra.

Dans la troisième section, nous présenterons les algorithmes développés pour construire des nuages de points colorés. Enfin, dans la dernière partie, nous montrerons ce que nous avons réalisé pour trianguler les nuages de points et comment nous créons les cartes de textures pour les modèles 3D texturés.

## 1. Construction de Modèles urbains 3D

### 1.1. Approches existantes

De nombreux travaux ont porté sur la modélisation d'environnements à partir de données terrestres. Nous pouvons différencier deux grandes approches pour la création de modèles : les systèmes basés sur des images et ceux basés sur le couplage de données image et laser.

Pour les systèmes basés sur des images, nous pouvons citer les travaux d'El Hakim [ELH02], qui présente des méthodes semi-automatiques de construction de modèles à partir de photos. Dick [DIC04] développe des méthodes entièrement automatiques de construction de modèles par reconnaissance bayésienne de sous-éléments architecturaux. Pénard et Paparoditis présentent dans [PEN05] un algorithme de génération automatique de surfaces texturées à partir d'un ensemble d'images calibrées. Ils ont développé pour cela un véhicule nommé Stéréopolis doté d'un ensemble de capteurs. Schlinder et Bauer [SCH03] proposent des méthodes de reconstruction de bâtiments à partir d'images pour obtenir un modèle de type CAO. Werner et Zisserman [WER02] présentent une stratégie de reconstruction de bâtiments à partir d'images non calibrées en générant des plans pour décrire la scène principale et en raffinant le modèle par des primitives géométriques. Enfin, Pollefeys construit, à partir de données images géo-référencées, des modèles 3D détaillés en temps réel dans [POL08]. Par cet aspect de temps réel, il se rapproche de nos travaux mais les images traitées de taille réduite (512x384 pixels) ont une résolution basse et un champ étroit.

Pour les systèmes basés scanner-laser et images, ceux-ci utilisent la grande précision du laser pour construire finement la géométrie et les images pour raffiner le modèle ou simplement comme texture. Bernardini pose dans [BER02] les problèmes fondamentaux que l'on rencontre dans toute la chaîne de construction de modèles 3D à partir de données images et laser : par exemple des problèmes de recalage, de maillage, de plaquage de textures… L'équipe de Tokyo avec Zhao et Shibasaki montre dans [ZHA03] un système complet de plate-forme mobile embarqué sur un véhicule pour la construction de modèles 3D mais la partie modélisation reste peu décrite. Stamos et Allen [STA06] présentent une approche de modélisation 3D qui modélise les zones planes par des primitives planes et les zones non planes par des surfaces triangulées. Früh et Zahor sont très proches de nos travaux dans [FRU05] où ils développent un ensemble d'algorithmes pour générer des façades texturées à partir de scans 2D verticaux et d'images.

Dans [STA06], les zones d'intérêt sont scannées de manière statique puis les nuages de points 3D sont fusionnés ensemble à partir des différentes localisations. Cela requiert bien sûr un traitement supplémentaire. Dans [FRU05, ZHA03], le véhicule est utilisé en tant que système de scanner : son mouvement définissant une des directions de l'environnement à modéliser. Dans ces trois dernières méthodes, le traitement des données collectées durant l'acquisition est fait à part et nécessite plusieurs heures de calculs ce qui ne permet pas d'avoir un système embarqué et temps réel.

### 1.2. Approche proposée

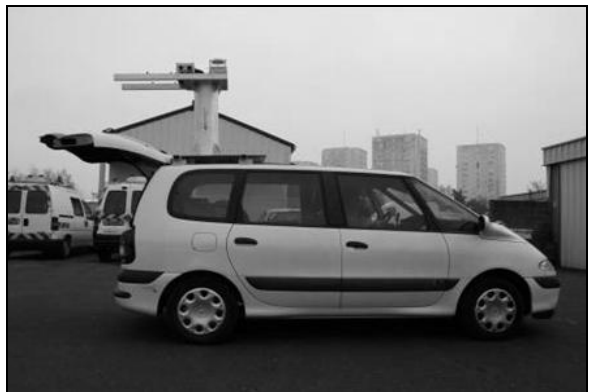

**Figure 1 Photo du prototype LARA-3D**

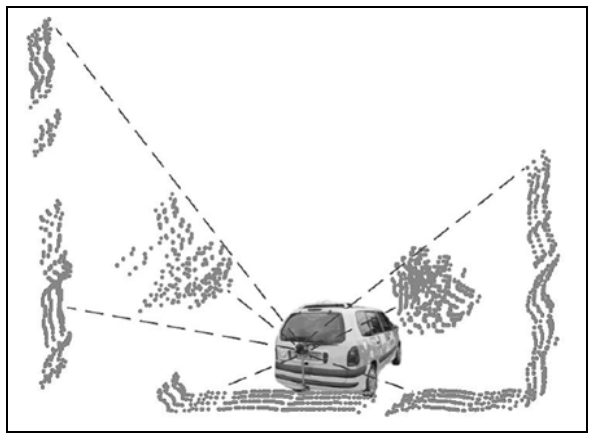

**Figure 2 Fonctionnement du scanner laser par profil**

Nous avons développé un système composé d'un véhicule avec une plate-forme équipée de capteurs de navigation (GPS, centrale inertielle) et d'un scanner laser. Le scanner laser, placé sur la plate-forme (Figure 1), balaie des plans verticaux perpendiculaires à la direction du véhicule (Figure 2).

Notre approche consiste à créer plusieurs types de modèles en temps réel à partir de notre système : des nuages de points 3D, des nuages de points 3D colorés, des modèles surfaciques et des modèles 3D surfaciques texturés. La Figure 3 présente toute la chaîne de traitement pour créer les différents types de modèles 3D.

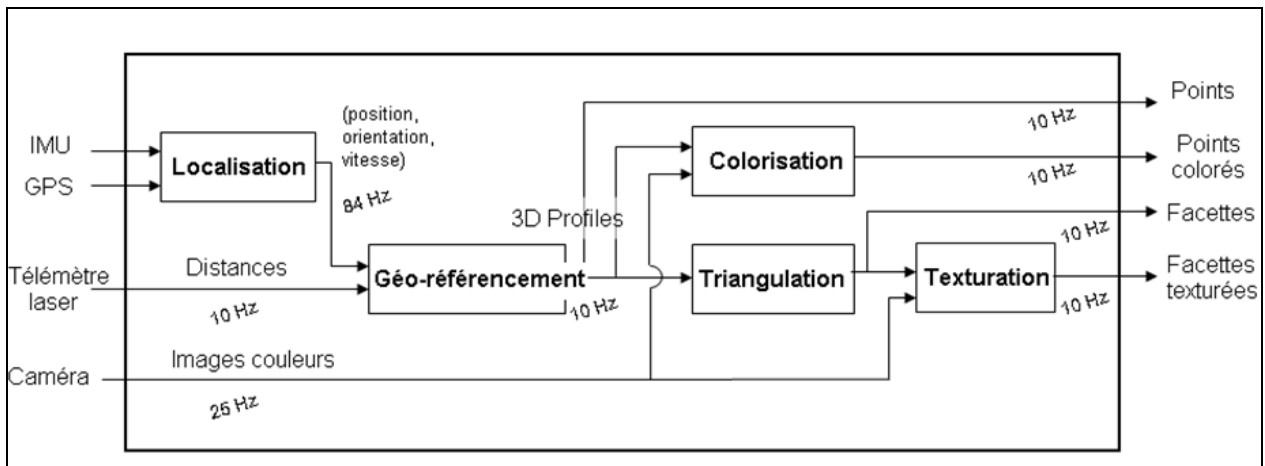

**Figure 3 Chaîne de traitement pour la création de modèles**

Nous pouvons nous poser la question du choix du temps réel. Nous voyons qu'avec la démocratisation des outils de cartographie mobile, il devient de plus en plus difficile de stocker les données brutes pour les traiter indépendamment. Dans notre cas, une estimation de la quantité de données brutes traitées a été faite et se porte à 2 Go de données par kilomètre. C'est pourquoi nous nous efforçons de concevoir un système entièrement embarqué qui permettrait d'obtenir les modèles texturés simplifiés sans avoir à stocker de données intermédiaires.

## 2. Choix matériels du nouveau prototype

### 2.1. Système existant

Notre prototype, appelé LARA-3D, est un véhicule équipé de capteurs de navigation (GPS, centrale inertielle), d'un scanner laser rotatif et d'un ordinateur embarqué avec un environnement logiciel adapté.

Pour la localisation, nous utilisons une antenne GPS Ag-GPS132 de Trimble, délivrant des positions absolues à une fréquence de 10 Hz et une centrale inertielle VG600CA-200 de Crossbow, délivrant des données à une fréquence de 84 Hz. Pour la localisation précise en temps réel, nous utilisons le couple GPS-centrale inertielle décrit dans [ANL03]. Le système de mesure inertielle donne seulement des informations dérivées (accélérations et vitesses de rotation) qui nécessitent d'être intégrées pour donner la position, l'orientation et la vitesse du véhicule. Ces informations à haute fréquence sont précises mais soumises à des dérives à long terme. L'information GPS donne des informations absolues (position et vitesse), mais à une faible fréquence et pas de façon régulière. La combinaison des deux, par fusion de données (filtre de Kalman), permet d'obtenir une position précise et absolue du véhicule ainsi que son orientation et sa vitesse.

Du point de vue logiciel, nous utilisons un environnement spécifique : une application temps réel embarquée nommée RTMAPS [Intempora]. RTMAPS permet d'implémenter des algorithmes de traitement en C++ et de les tester sur des données réelles déjà acquises, c'est à dire de faire rejouer les données à des vitesses plus faibles ou plus grandes que lors de l'acquisition originale. Il permet surtout de donner un temps précis sur chaque donnée ce qui permet de synchroniser les données des différents capteurs lors de leur fusion. L'ordinateur embarqué utilisé pour acquérir toutes les données sous RT-MAPS est un AEC-6920 avec un processeur Core 2 Duo.

Le scanner laser est une LD Automotive d'IBEO, couvrant un angle de 270° et délivrant 1080 distances avec une fréquence de rotation de 10 Hz. Il fournit des données sous forme de distance et d'angle dans un plan mobile avec le véhicule. Nous obtenons des profils (c'est-à-dire des données 2D) comme sur la Figure 4.

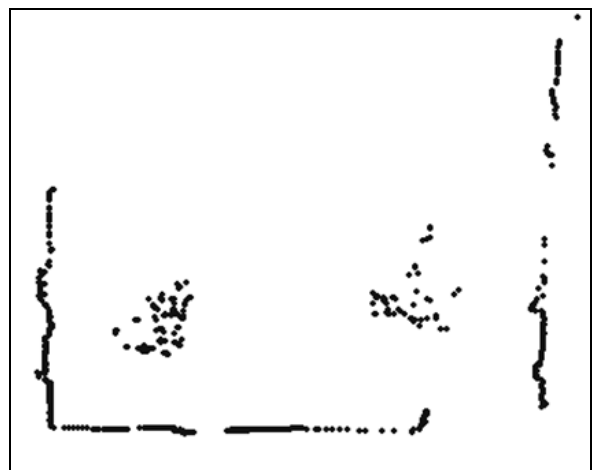

**Figure 4 Exemple d'un profil du scanner laser**

Avec l'information de localisation de la voiture et les données du scanner laser, on transforme ces points 2D situés dans un plan mobile en points 3D dans un référentiel commun (on néglige ici la faible divergence du faisceau du laser pour considérer des points géométriques). Nous produisons alors un nuage de points 3D dans un référentiel commun terrestre (en Lambert II) (Figure 5).

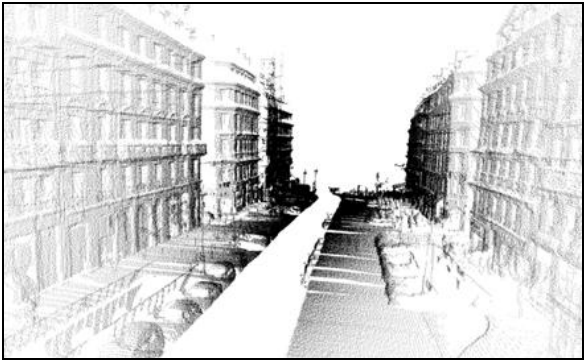
**Figure 5 Nuage de points 3D**

La précision des points du nuage est fortement dépendante de la précision de la localisation. A partir d'une série de tests de modélisation d'une route, nous avons mesuré la précision de nos données avec un écart-type de 5 cm et un biais inférieur à 4 m. Il faut ajouter à cela un bruit de mesure sur la capture d'un point avec le scanner laser qui est de l'ordre de 5 cm [GOU06]. La précision de localisation a un impact important sur la colorisation et la texturation. Cela sera discuté par la suite.

**2.2. Caméra et optique choisis**

Notre objectif a été d'ajouter une information de couleur à ces nuages de points 3D. Nous avons considéré de nombreuses possibilités.

Nous avons fait une comparaison des angles couverts par une caméra avec trois objectifs différents et nous avons finalement choisi d'utiliser une caméra avec un objectif fish-eye pour une capture très grand angle. Les objectifs fish-eye sont des lentilles spécifiques utilisant des principes optiques différents des lentilles classiques, qui permettent d'obtenir des longueurs de focales très courtes et ainsi de capturer la moitié de l'espace avec une seule caméra. Pour couvrir la même zone que le scanner, seulement deux caméras avec objectif fish-eye sont nécessaires, alors qu'un objectif classique aurait nécessité au moins quatre caméras. Le but était d'utiliser le moins de caméras possible pour des raisons de complexité technique et de coût. Nous avons aussi envisagé d'utiliser des caméras panoramiques mais les objectifs fish-eye sont plus disponibles en terme de produit et ont une optique plus facilement modélisable.

Pour avoir une redondance photogrammétrique (chevauchement des images), nous avons choisi une caméra CCD matricielle pour obtenir la scène complète à chaque image (différent d'une caméra CCD linéaire).

Tout ceci explique notre choix d'une caméra CCD avec objectif fish-eye pour les images. La caméra a été fixée rigidement sur la même plate-forme que le scanner laser (Figure 6).

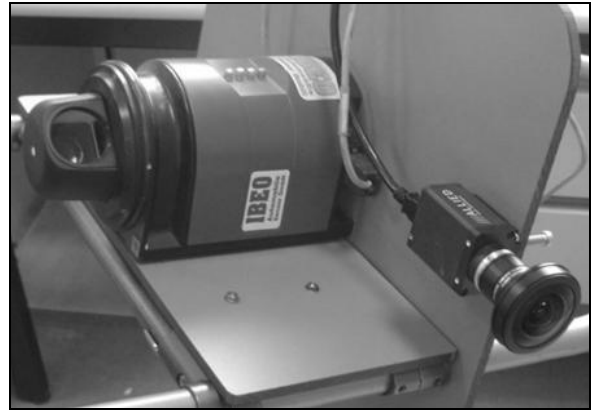
**Figure 6 Plate-forme rigide scanner laser/caméra**

Les premiers résultats ont été obtenus avec une caméra Marlin F046C avec une résolution de 776x580 et une fréquence de 30 Hz (Figure 7). Nous sommes ensuite passés à une caméra plus haute résolution Pike F421C de résolution 2048x2048 à une fréquence de 1 Hz (Figure 8).

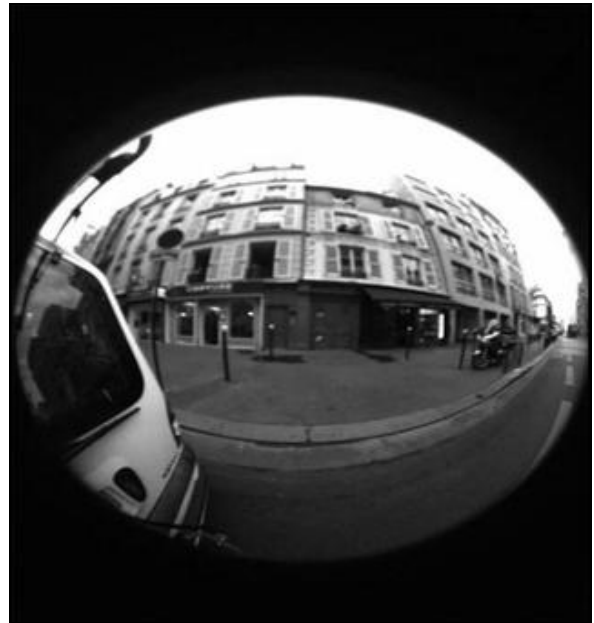
**Figure 7 Image de la caméra Marlin avec objectif fish-eye**

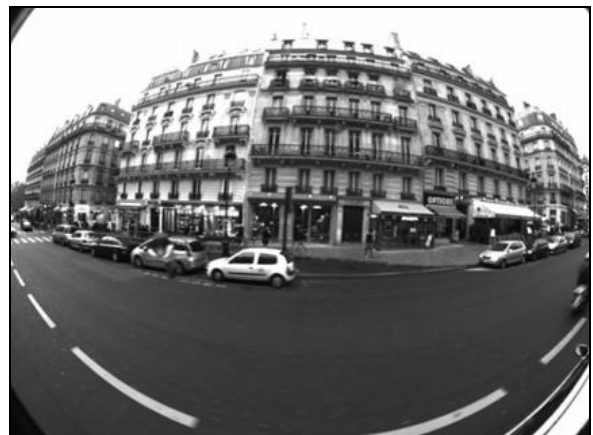
**Figure 8 Image de la caméra Pike avec objectif fish-eye**

## 2.3 Etalonnage scanner laser/caméra

Pour utiliser les données de la caméra et du scanner laser dans le même référentiel, nous avons besoin de connaître les transformations géométriques pour passer du référentiel laser au référentiel caméra. Cette transformation est rigide et doit être déterminée de façon précise par un étalonnage. Nous avons développé une méthode détaillée dans [DKF05, BRU07c].

L'idée de base est d'utiliser une mire d'étalonnage plane sous forme d'échiquier. Les points du laser doivent tomber sur la mire plane dont la position est estimée par les images de la caméra. Nous avons alors une contrainte géométrique sur la transformation rigide entre le système de coordonnées de la caméra et le système de coordonnées du laser. Nous effectuons les mesures de chaque capteur au moins une quinzaine de fois et à chaque fois, la position de la mire est différente. Pour ce processus d'étalonnage, nous utilisons une lentille "classique" car nous avons juste besoin des paramètres extrinsèques de la caméra.

## 2.4. Etalonnage du Fish-eye

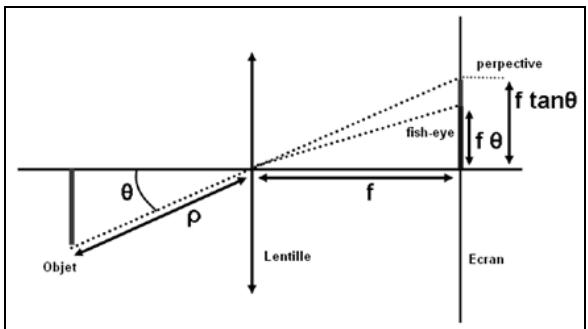

**Figure 9 Différence entre une projection perspective une projection équidistante**

Nous avons équipé la caméra d'un objectif fish-eye dans le but d'avoir un grand angle d'ouverture (Figure 7 et Figure 8). En conséquence, le modèle standard de projection perspective ne peut plus être utilisé, nous avons besoin d'un autre modèle de projection. Nous avons décidé d'utiliser le modèle équidistant (Figure 9) qui est une bonne approximation du comportement de projection d'une lentille fish-eye [BRU07c] et plus précisément, nous utilisons une généralisation du modèle équidistant définie dans [KAN06]. Ce dernier article propose trois modèles de caméras à 6, 9 ou 23 paramètres. Le dernier modèle à 23 paramètres tient compte des distorsions radiales et tangentielles. En calibrant la caméra Pike F421C, nous avons obtenu une erreur résiduelle moyenne de 0,50 pixel pour le modèle à 6 paramètres, 0,46 pixel pour le modèle à 9 paramètres et 0,44 pixel pour celui à 23 paramètres. Nous avons choisi le meilleur compromis entre simplicité du modèle et qualité de la reprojection. Entre le modèle à 6 paramètres et à 9 paramètres, il y a une amélioration de 9% sur l'erreur résiduelle moyenne alors que l'on note seulement une amélioration de 4% entre le modèle à 9 paramètres et celui à 23. Nous avons donc choisi le modèle de projection à 9 paramètres pour garder une bonne qualité de la projection et une certaine rapidité de traitement.

L'étalonnage correct de la caméra nous permet de transformer les images fish-eye en perspective ou en ortho-photo (Figure 10). Cette correction est nécessaire pour construire les cartes de textures corrigées.

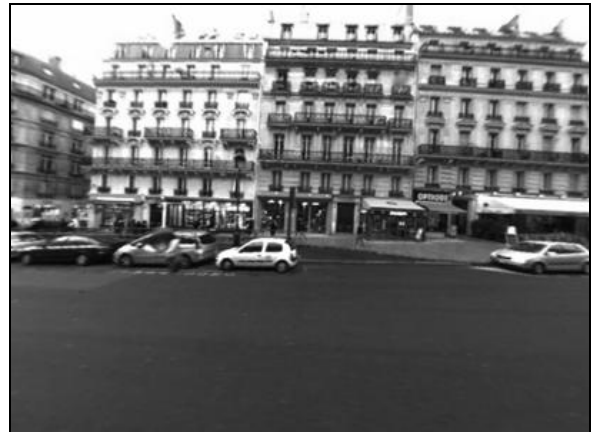

**Figure 10 Correction de l'image en vue perspective**

Tous ces étalonnages se font hors acquisitions et nous considérons qu'ils restent valables lors des acquisitions : nous négligeons les effets des vibrations dues au moteur et au mouvement du véhicule.

### 3. Production de points 3D colorés

Pour obtenir la couleur de chaque point du nuage de points 3D, nous avons développé deux algorithmes qui dépendent de la fréquence d'acquisition des images.

Pour préciser les ordres de grandeur, nous avons environ 1000 points dans chaque profil tous les 100ms avec le scanner laser, 30 images par seconde pour la caméra Marlin F046C (résolution de 776x580) et une image par seconde pour la caméra Pike F421C (résolution 2048x2048). Le véhicule se déplace à une vitesse lente (environ 5 km/h) pour avoir une plus grande résolution horizontale. Cela représente un espace d'environ 15 cm entre deux profils adjacents. Avec une vitesse de 5 km/h, un point fixe d'une façade se sera déplacé en 100 ms d'environ 15 cm par rapport au repère mobile de la voiture (en considérant le déplacement de la voiture comme parallèle à la façade).

### 3.1. Colorisation par couplage direct profil laser-image

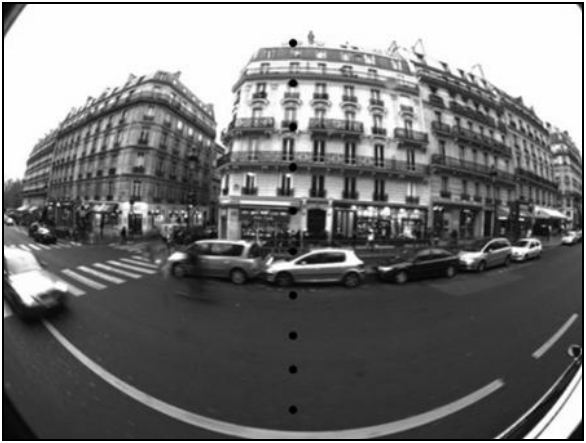
**Figure 11 Couplage scanner laser/image**

Avec une caméra de fréquence suffisante (30 Hz pour une Marlin F046C), il est possible d'associer à chaque profil laser une image de la caméra dont le temps d'acquisition est suffisamment proche du temps d'acquisition du profil laser. On raisonne alors par couple « profil laser-image ». Le principe est de considérer le couplage mécanique de la plate-forme constituée du scanner et de la caméra comme rigide. On synchronise ensuite chaque donnée scanner avec une donnée de la caméra.

Plus précisément, quand on reçoit une trame de données du scanner laser, cela correspond à 1080 points (un profil) qui sont datés par notre application RT-MAPS de synchronisation de données. Les données images sont aussi datées. On sélectionne parmi les dernières images reçues de la caméra l'image acquise au moment le plus proche de la date d'acquisition du profil laser. On obtient un couple « profil laser-image ». On utilise alors les paramètres de la transformation rigide (déterminés lors de l'étape d'étalonnage scanner/caméra décrite dans la section 2) pour passer les points du repère laser dans le repère caméra. Dans le repère caméra, les points sont projetés à l'aide du modèle de projection dont les paramètres ont été définis à l'étape d'étalonnage du fish-eye. Nous obtenons alors des points laser dans l'image fish-eye comme sur la Figure 11. A chaque point laser, on attribue une couleur qui permet de construire un nuage de points 3D coloré. Les résultats de cette colorisation sont sur la Figure 16.

Nous voyons que pour avoir une précision suffisante, il faut que la différence de temps d'acquisition entre une image et un profil laser soit la plus petite possible. Il est donc nécessaire d'avoir une fréquence d'image suffisante pour que les images soient acquises à un temps suffisamment proche du profil laser. Nous cherchons à connaître la différence de temps d'acquisition entre un profil et une image. Avec une fréquence du scanner de 10 Hz (donc un profil tous les 100 ms) et une caméra à une fréquence de 30 Hz, cette différence sera de +/- 16 ms, ce qui fait en valeur moyenne absolue un écart de 8 ms. Sachant que le véhicule se déplace à 5 km/h, un point fixe sur la façade aura bougé d'environ 1,2 cm dans le repère mobile de la caméra pendant les 8 ms. Si l'on considère ce point comme étant à une distance de 5 m de la caméra, cela va représenter une erreur de projection de 0,4 pixel en moyenne (0,8 pixels au pire) sur une image de la caméra Marlin. Cette erreur de projection due à la synchronisation peut donc être négligée.

### 3.2. Colorisation par mise en correspondance de données images et laser géoréférencées

Nos premiers tests ont été réalisés à l'aide de la caméra basse résolution et à haute fréquence. En passant à une caméra haute résolution et basse fréquence (1 Hz dans notre cas), nous ne pouvons plus utiliser de couplage scanner laser/caméra comme décrit précédemment. La faible fréquence induit un manque d'images qui ne permet plus de faire de la synchronisation avec les données laser : nous avons une erreur moyenne de 250 ms (500 ms au pire) pour la synchronisation scanner/laser. En utilisant le même raisonnement développé dans le paragraphe précédent, 250 ms d'erreur va représenter un déplacement d'environ 37 cm pour un point fixe d'une façade par rapport au repère mobile du véhicule. Si on projette cette erreur sur une image de la caméra Pike, cela fait une erreur de 23 pixels en moyenne (46 pixels au pire) ce qui n'est pas du tout exploitable.

Nous avons donc considéré les données image comme une source de données indépendante. Avec la localisation précise du véhicule (à une fréquence de 10 Hz et une date « d'acquisition » pour chaque position qui va être calculée par l'algorithme de localisation), nous pouvons géolocaliser chaque image de la caméra dans le même référentiel que celui du nuage de points 3D. A chaque image reçue de la caméra, on trouve la position du véhicule dont la date d'acquisition est la plus proche de la date d'acquisition de l'image. Avec une localisation à 10 Hz et une fréquence d'image à 1 Hz, cela donne une erreur moyenne de synchronisation de 25 ms (50 ms au pire). Pour une erreur de 25 ms, nous obtenons avec le même raisonnement développé plus haut une erreur de projection d'environ 2 pixels en moyenne (4 pixels au pire).

Nous avons amélioré cette erreur de projection due à la synchronisation localisation/image en utilisant des régressions linéaires de la position et de l'orientation entre deux points de la trajectoire. Plus précisément, pour chaque image, on trouve les deux positions dont les dates d'acquisitions encadrent la date d'acquisition de l'image. On estime ensuite par régression linéaire entre ces deux positions, la position dont la date d'acquisition correspond à la date d'acquisition de l'image. Cette étape permet de diminuer l'erreur de projection due à la synchronisation mais cette amélioration reste difficile à quantifier.

Dans une deuxième étape, pour chaque point 3D du nuage de points dans le même repère terrestre, nous récupérons l'image géolocalisée la plus proche géométriquement. Nous calculons pour cela la distance entre le point du nuage et le centre des caméras pour chaque image. Une fois l'image la plus proche sélectionnée, nous pouvons passer le point 3D du repère terrestre dans le repère de la caméra puis projeter le point dans l'image à l'aide du même modèle de projection et récupérer sa couleur. Le nuage de points coloré avec la caméra haute résolution est visible sur la Figure 17.

Tous ces traitements se font en « temps réel différé ». C'est-à-dire que nous stockons les données (position, orientation, laser et caméra) dans une mémoire tampon (de l'ordre de quelques secondes) de manière à avoir suffisamment de données pour faire les traitements. Cela signifie que le nuage de points 3D coloré sera obtenu quelques secondes après l'arrêt des acquisitions à bord du véhicule, le temps de traiter la fin des données. Cela s'explique par le fait qu'avec une fréquence de 1 Hz pour la caméra, nous pouvons acquérir des profils laser dont l'image correspondante n'a pas encore été acquise. Il faut donc stocker toutes ces données temporairement avant de les traiter.

### 3.3. Comparaison

Nous venons de donner deux méthodes pour coloriser un nuage de points à partir d'une caméra.

Pour ces deux méthodes, la précision colorimétrique de la colorisation comme la précision géométrique du nuage de points 3D vont dépendre fortement de la précision des étalonnages présentés plus haut (étalonnage caméra et scanner laser/caméra) mais aussi de la précision de la localisation (c'est-à-dire position et orientation). Par exemple, une erreur de 1° sur l'estimation du tangage va entraîner une erreur de 15 cm sur la position du point du scanner laser. Cette précision va donc aussi se répercuter sur la précision colorimétrique de la colorisation.

La première méthode de couplage fort va nous permettre d'obtenir des points colorés avec une plus grande précision colorimétrique. En effet, la précision dépendra uniquement de la précision des étalonnages et de la synchronisation entre un profil et l'image acquise la plus proche. Cela nécessite des données fortement couplées donc une fréquence d'acquisition des images suffisamment grande (donc des images de plus basse résolution).

Pour la deuxième méthode, l'avantage est de pouvoir utiliser des caméras haute résolution, de permettre aussi l'utilisation de données acquises indépendamment (on pourrait imaginer utiliser d'autres images géoréférencées de meilleures résolutions pour les textures) mais la précision va dépendre de la précision du géoréférencement des deux sources. Même en utilisant un géoréférencement identique pour les deux sources de données (images et laser) comme cela a été le cas dans notre expérimentation, la précision sera plus grossière que dans le cas du couplage fort.

L'avantage de ces deux méthodes est de traiter les données en temps réel ou « temps réel différé » ce qui permet d'obtenir un nuage de points 3D coloré directement à la fin des acquisitions.

### 4. Production de modèles texturés

Comme nous l'avons vu sur la Figure 3, notre chaîne de traitement nous permet de créer des nuages de points 3D et des nuages de points colorés avec l'étape de colorisation vue dans la section précédente. Mais nous pouvons aussi créer des modèles surfaciques et des modèles surfaciques texturés en temps réel. Nous allons voir l'étape de triangulation et de texturation dans cette partie.

#### 4.1. Triangulation temps réel

Les méthodes de triangulation de nuages de points 3D sont nombreuses dans la littérature mais elles sont en général incompatibles avec une exécution temps réel. Comme cela est décrit dans [BER02], il existe 4 grandes approches pour la triangulation : les méthodes basées sur la triangulation de Delaunay, celles basées sur la création d'une surface par une paramétrisation locale, celles volumétriques comme les « marching cubes » et enfin celles basées sur les surfaces déformables. Ce sont surtout les méthodes locales et volumétriques qui sont utilisables pour trianguler de grandes quantités de données en des temps suffisamment courts.

Nous avons pour l'instant voulu développer une méthode simple temps réel de triangulation qui sera améliorée par la suite. Nous utilisons pour cela les types de données dont nous disposons. Ces données sont fournies en 2D sous forme de profil (Figure 4) qui représente une découpe de la rue perpendiculaire à la direction du véhicule. Un profil est construit par une donnée d'angle de mesure θ et de distance r.

La triangulation s'effectue entre deux profils adjacents. Entre deux profils successifs, les points ayant le même angle de mesure θ se retrouvent voisins. Plus précisément, si $\theta_i$ et $\theta_{i+1}$ sont deux points d'angles voisins dans un même profil et $\theta_i'$ et $\theta_{i+1}'$ les deux points de même angle dans le profil suivant, alors on va créer les triangles ($\theta_i$, $\theta_i'$, $\theta_{i+1}'$) et ($\theta_i$, $\theta_{i+1}$, $\theta_{i+1}'$). Nous voyons le résultat de la triangulation entre deux profils sur la Figure 12. On peut ainsi répéter l'opération pour tous les profils et on obtient le résultat pour une scène urbaine visible sur la Figure 13.

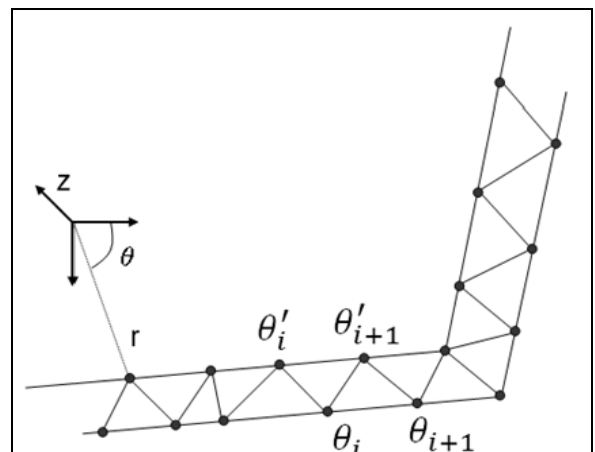

**Figure 12 Triangulation entre deux profils**

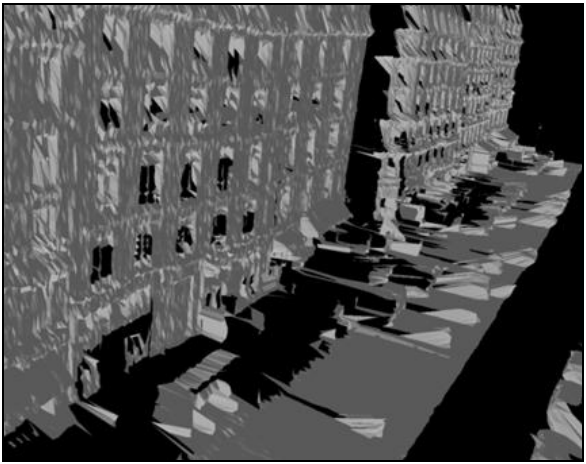

**Figure 13 Modèle 3D triangulé**

### 4.2. Texturation temps réelle

#### 4.2.1. Calcul des cartes de textures

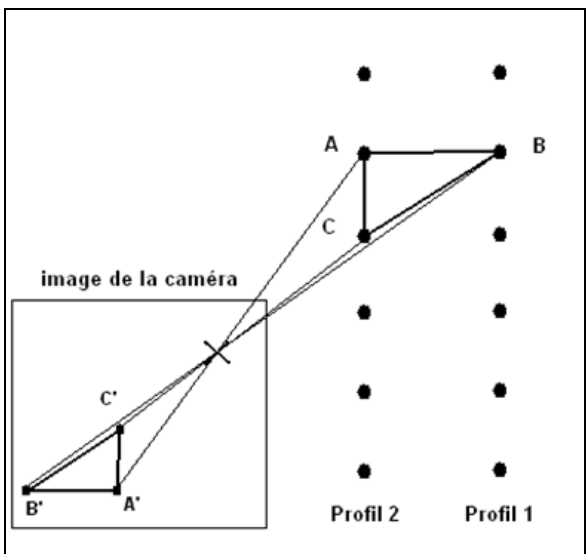

**Figure 14 Illustration de la projection des triangles sur les images de la caméra**

A partir du modèle 3D triangulé, nous pouvons ajouter des textures en utilisant les images de la caméra.

Le même algorithme itératif est appliqué pour chaque triangle. Un triangle est défini par trois points et d'après l'algorithme de triangulation décrit dans la partie précédente, deux points sont dans le même profil et un point appartient au profil adjacent. On va donc choisir une image dont la date d'acquisition est la plus proche d'un des deux profils. Pour ce profil là, on projette le ou les points du triangle du repère laser dans l'image correspondante. Pour le profil adjacent, on détermine tout d'abord le ou les points dans le repère monde, puis avec la géolocalisation de la caméra, on peut calculer de nouveau ces coordonnées dans le repère caméra. On obtient donc les 3 points du triangle dans le même repère de la caméra.

A partir des 3 points dans le repère caméra, on peut projeter les 3 points de chaque triangle sur l'image fish-eye à l'aide du modèle de projection défini lors de l'étape d'étalonnage (Figure 14). On obtient ainsi une texture sous forme de triangle que l'on peut plaquer sur le triangle choisi au début de l'algorithme.

Après cela, nous enregistrons les textures dans des cartes de textures. Nous entourons les triangles images par des rectangles images (boîte englobante) et les enregistrons côte à côte dans des fichiers BMP. OpenGL peut alors utiliser les cartes de textures et les modèles surfaciques pour afficher les modèles 3D texturés. Ce travail a été effectué pour la caméra basse-résolution et un résultat est visible sur la Figure 18.

Tous ces traitements sont effectués en « temps réel différé » (les calculs se font à la même vitesse que l'acquisition des données mais nous devons garder un délai de quelques profils pour le calcul de la triangulation et de la texturation).

#### 4.2.2. Optimisation des cartes de textures

Nous enregistrons les textures dans des cartes de textures (fichiers BMP) en entourant les triangles par des rectangles (boîte englobante). Pour gagner de la mémoire (ceci est important pour afficher les modèles plus rapidement), nous avons développé un algorithme pour diminuer le nombre de cartes de textures. Pour cela, nous effectuons une classification des rectangles en les regroupant suivant leur hauteur [RCM99]. Par exemple, il y a une carte de texture avec toutes les textures de triangles dont la hauteur de la boite englobante est comprise entre 2 et 5 pixels. Il y a une autre carte de texture pour les textures dont la hauteur de la boite englobante est comprise entre 6 et 8 pixels… Avec cette classification, nous sommes passés de 302 cartes de textures à seulement 108 cartes pour le modèle d'une rue (Figure 15).

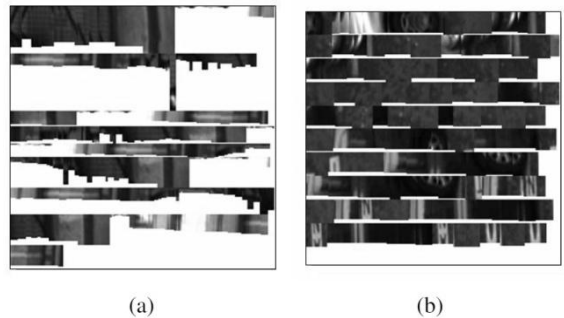

**Figure 15 Amélioration des cartes de textures**

### 4.3. Analyse des résultats

La méthode de triangulation et de texturation que nous venons de décrire nous permet de construire des modèles texturés en temps réel.

Comme pour la création des nuages de points colorés, la précision de la localisation est un facteur important pour la précision des modèles 3D texturés.

Mais ces modèles 3D ont encore certaines limites : tout d'abord, notre triangulation ne permet pas de tenir compte des objets sémantiques de la scène. En reliant deux points de même angle de mesure θ entre deux profils adjacents avec le scanner laser, nous pouvons relier un point de la façade avec une voiture.

Nous remarquons aussi des bavures des textures sur les hauts des immeubles (voir la Figure 18). Ceci est principalement causé par la faible résolution de la caméra Marlin F046C qui est de 400 000 pixels (10cm sur la façade d'un immeuble à 5 m de distance représentent 3 pixels sur l'image).

**Conclusion**

Nous avons présenté deux méthodes de colorisation de nuage de points et une méthode de construction de modèle texturés, tout ceci en temps réel ou temps réel différé. Ces résultats sont basés sur le couplage et l'étalonnage de deux capteurs, un scanner laser et une caméra équipée d'un objectif fish-eye. Les résultats montrent que la texturation temps réel est possible. Ces différents travaux ont aussi permis de valider les différentes étapes d'étalonnage. Mais pour tester de façon fiable ces résultats, il y aurait un vrai besoin de mettre en place des méthodes permettant de mesurer précisément les nuages de points 3D colorés et les modèles 3D.

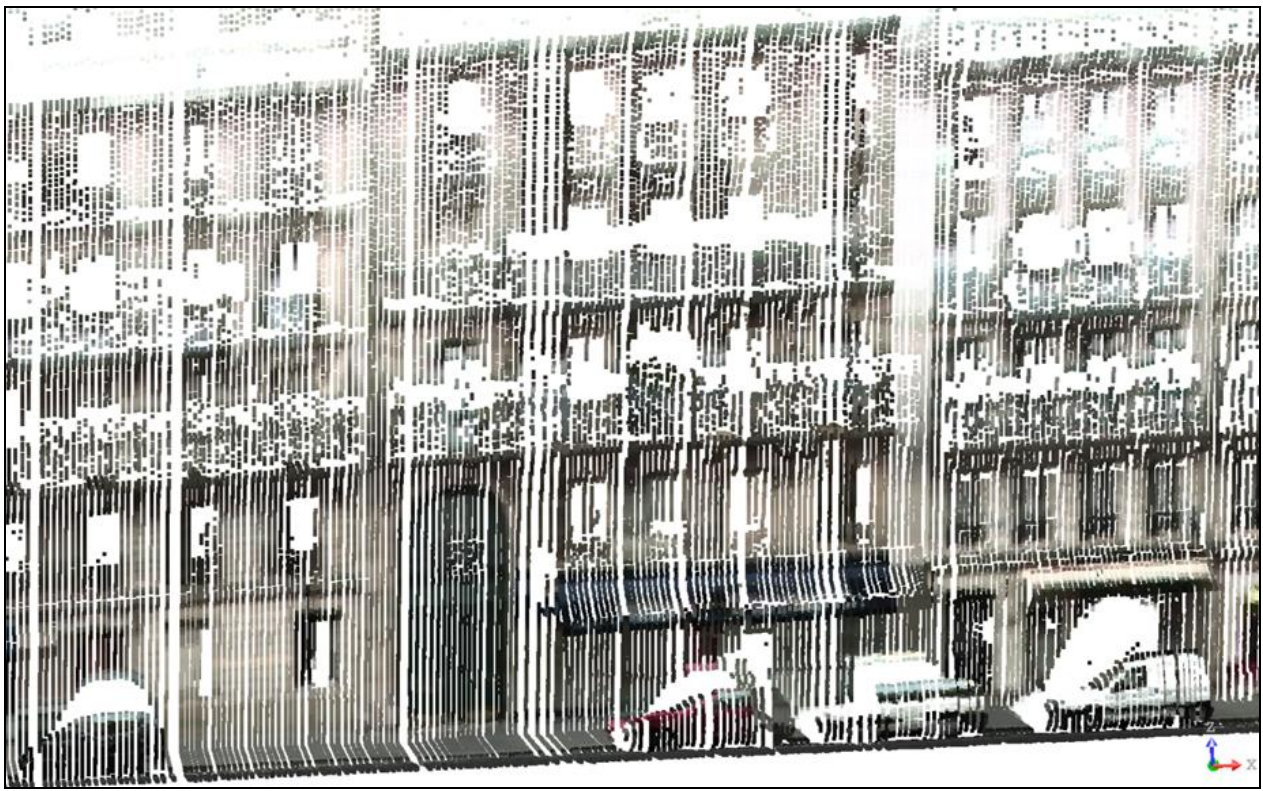

**Figure 16 Nuage de points colorés avec caméra basse résolution**

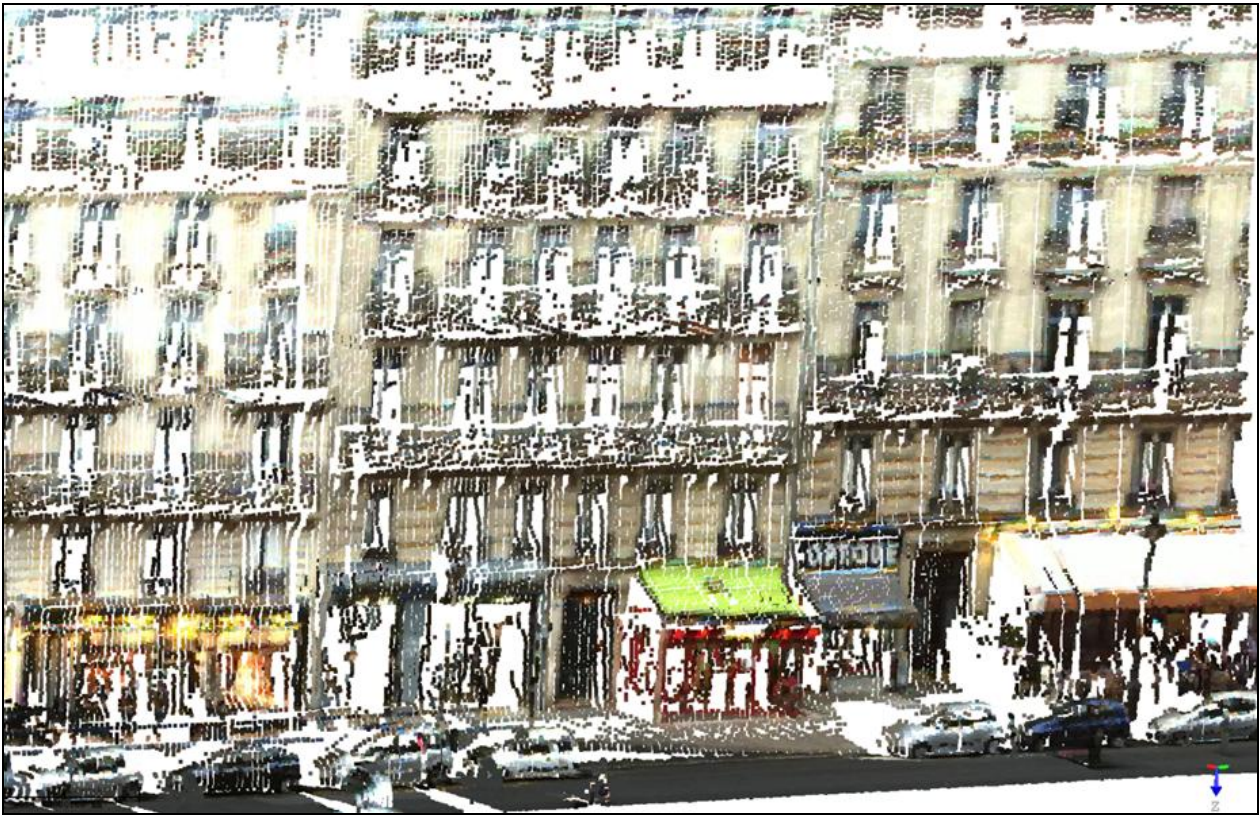

**Figure 17 Nuage de points colorés avec caméra haute résolution**

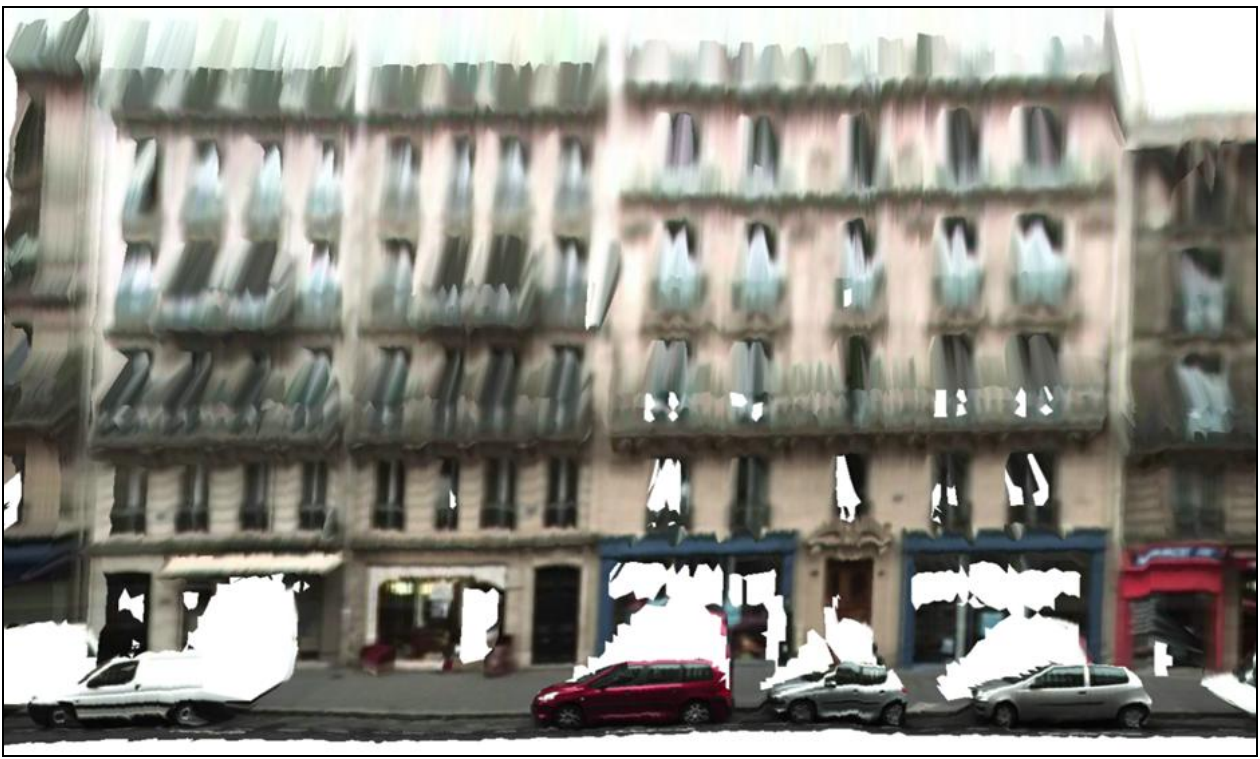

**Figure 18 Modèle 3D texturé avec caméra basse résolution**

*Revue Française de Photogrammétrie et de Télédétection*